\documentclass{article} 
\usepackage{iclr2016_conference,times}
\usepackage{hyperref}
\usepackage{url}

\usepackage{mathtools}
\usepackage{amsmath}
\usepackage{amsfonts}

\title{
Recurrent Models for \\
Auditory Attention in Multi-Microphone \\
Distance Speech Recognition}

\author{Suyoun Kim\\
Electrical Computer Engineering\\
Carnegie Mellon University\\
\texttt{suyoun@cmu.edu}\\
\And
Ian Lane\\
Electrical Computer Engineering\\
Carnegie Mellon University\\
\texttt{lane@cmu.edu}
}

\begin{document}

\maketitle

\begin{abstract}

Integration of multiple microphone data is one of the key ways to achieve robust speech recognition in noisy environments or when the speaker is located at some distance from the input device. Signal processing techniques such as beamforming are widely used to extract a speech signal of interest from background noise. These techniques, however, are highly dependent on prior spatial information about the microphones and the environment in which the system is being used. In this work, we present a neural attention network that directly combines multi-channel audio to generate phonetic states without requiring any prior knowledge of the microphone layout or any explicit signal preprocessing for speech enhancement. We embed an attention mechanism within a Recurrent Neural Network (RNN) based acoustic model to automatically tune its attention to a more reliable input source. Unlike traditional multi-channel preprocessing, our system can be optimized towards the desired output in one step. Although attention-based models have recently achieved impressive results on sequence-to-sequence learning, no attention mechanisms have previously been applied to learn potentially asynchronous and non-stationary multiple inputs. We evaluate our neural attention model on the CHiME-3 challenge task, and show that the model achieves comparable performance to beamforming using a purely data-driven method. 


\end{abstract}

\section{Introduction}
\label{sec:intro}

Many real-world speech recognition applications, including teleconferencing, robotics and in-car spoken dialog systems, must deal with speech from distant microphones in noisy environments. When a human voice is captured with far-field microphones in these environments, the audio signal is severely degraded by reverberation and background noise. This makes the distant speech recognition task far more challenging than near-field speech recognition, which is commonly used for voice-based interaction today.

Acoustic signals from multiple microphones can be used to enhance recognition accuracy due to the availability of additional spatial information. Many researchers have proposed techniques to efficiently integrate inputs from multiple distant microphones. The most representative multi-channel processing technique is the beamforming approach \citep{van1990speech, seltzer2004likelihood, kumatani2012microphone, pertila2015distant}, which generates an enhanced single output signal by aligning multiple signals through digital delays that compensate for the different distances of the input signals. However, the performance of beamforming is highly dependant on prior information about microphone location and the location of the target source. For downstream tasks such as speech recognition, this preprocessing step is suboptimal because it is not directly optimized towards the final objective of interest: speech recognition accuracy \citep{seltzer2008bridging}. 

Over the past few years, deep neural networks (DNNs) have been successfully applied to acoustic models in speech recognition \citep{seide2011conversational, mohamed2012acoustic, hinton2012deep}. Other works \citep{liu2014using, renals2014neural, swietojanski2014convolutional, yoshioka2015far, himawan2015learning} have shown that DNNs can learn suitable representations for distant speech recognition by directly using multi-channel input. These approaches however, simply concatenated acoustic features from multiple microphones without considering the spatial properties of acoustic signal propagation, or used convolutional neural networks (CNNs) to implicitly account for spatial relationships between channels \citep{renals2014neural, swietojanski2014convolutional}.

Recently, an "attention mechanism" in neural networks has been proposed to address the problem of learning variable-length input and output sequences \citep{bahdanau2014neural}. At each output step, the previous output history is used to generate an attention vector over the input sequence. This attention vector enables models to learn to focus attention on specific parts of their input. These attention-equipped frameworks have shown very promising results on many challenging tasks involving inputs and outputs with variable length, including machine translation \citep{bahdanau2014neural}, parsing \citep{vinyals2014grammar}, image captioning \citep{xu2015show} and conversational modeling \citep{vinyals2015neural}. Specifically, for the speech recognition tasks, \citep{chorowski2014end, chan2015listen, bahdanau2015end} attempted to align the input features and the desired character sequence using an attention mechanism. However, no attention mechanisms have been applied to learn to integrate multiple inputs.

In this work, we propose a novel attention-based model that enables to learn misaligned and non-stationary multiple input sources for distant speech recognition. We embed an attention mechanism within a Recurrent Neural Network (RNN) based acoustic model to automatically tune its attention to a more reliable input source among misaligned and non-stationary input sources at each output step. The attention module is learned with the normal acoustic model and is jointly optimized towards phonetic state accuracy. Our attention module is unique in the way that we 1) deal with the problem of integrating different qualities and misalignment of multiple sources, and 2) exploit spatial information between multiple sources to accelerate learning of auditory attention. Our system plays a similar role to traditional multichannel preprocessing through deep neural network architecture, but bypasses the limitations of preprocessing, which requires an expensive, separate step and depends on prior information. 

Through a series of experiments on the CHiME-3 \citep{chime3} dataset, we show that our proposed approach improves recognition accuracy in various types of noisy environments. In addition, we also compare our approach with the beamforming technique\citep{chime3,loesch2010adaptive,blandin2012multi,mestre2003diagonal}. The paper is organized as follows: in Section \ref{sec:model} we describe our proposed attention based model. In section \ref{sec:exp}, we evaluate the performance of our model. Finally, in Section \ref{sec:conclusion} we draw conclusions.

\section{Model}
\label{sec:model}
In this section, we describe our neural attention model, which allows neural networks to focus more on reliable input sources across different temporal locations. We formulate the proposed framework with applications in multi-channel distant speech recognition. While there has been some recent work on end-to-end neural speech recognition systems - from speech directly to transcripts \citep{graves2006connectionist, graves2014towards, hannun2014deepspeech, chorowski2014end} - our model is based on typical hybrid DNN-HMM frameworks \citep{morgan1994connectionist, hinton2012deep}, wherein the acoustic model estimates hidden Markov model (HMM) state posteriors, because we focus on dealing with the re-weighted input representation of misaligned multiple input sources. 

Given a set of input sequences $\mathbf X = \{ \mathbf X^{ch_1}, \cdots, \mathbf X^{ch_N} \}$, where $\mathbf X^{ch_i}$ is an input sequence $({x^{ch_i}_1}, \cdots, {x^{ch_i}_T})$ from the $i$th microphone, $i \in \{1, \cdots, N\}$, our system computes a corresponding sequence of HMM acoustic states, $\mathbf y = (y_1,\cdots, y_T)$. We model each output $\mathbf y_t$ at time $t$ as a conditional distribution over the previous outputs $y_{<t}$ and the multiple inputs $\mathbf X_t$ at time $t$ using the chain rule:

\begin{equation}
P(\mathbf y|\mathbf X) = \prod_t P(y_t | \mathbf X, y_{<t})
\end{equation}

\begin{figure}[h]
\begin{center}
   \includegraphics[width=\linewidth]{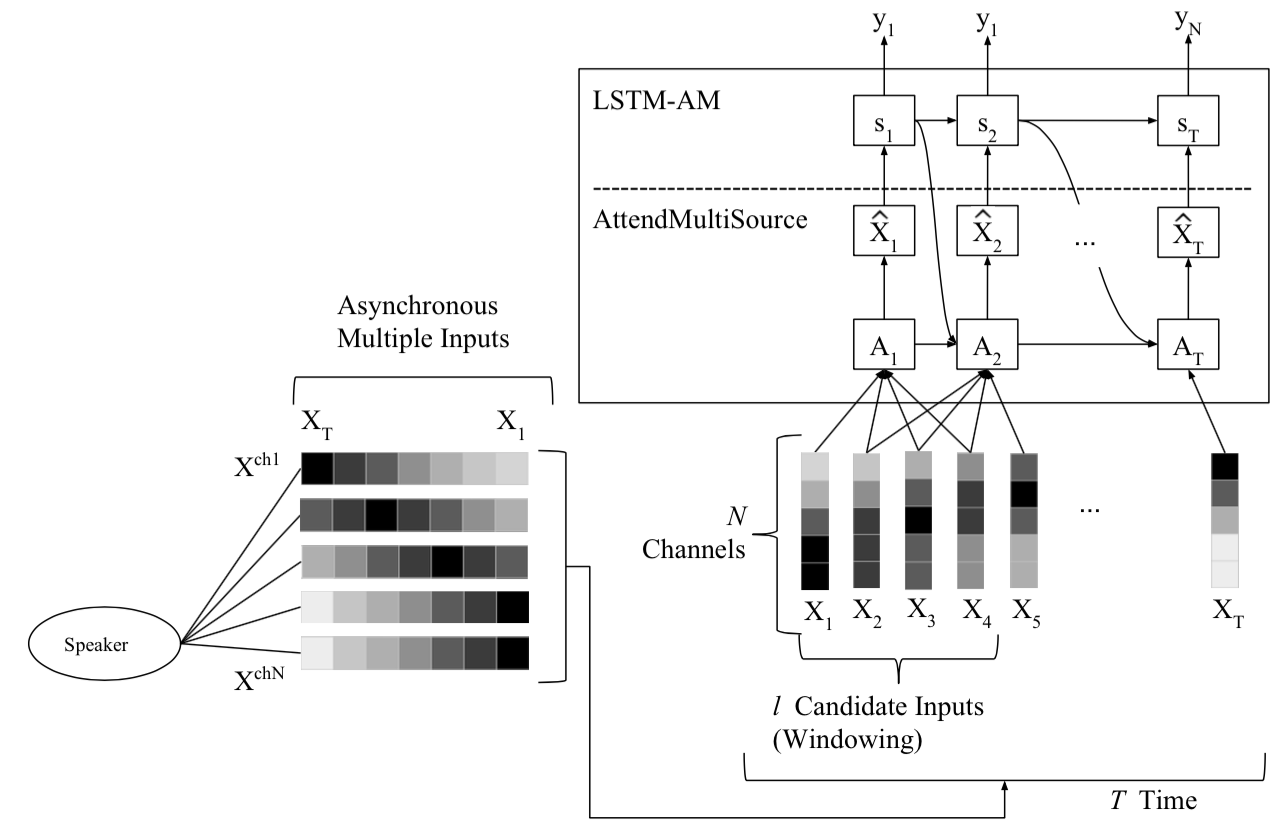}
\end{center}
\caption{Schematic representation of our neural attention model.}
\label{fig:schematic}
\end{figure}

Our system consists of two subnetworks: $\operatorname{AttendMultiSource}$
and $\operatorname{LSTM-AM}$. $\operatorname{AttendMultiSource}$ is an attention-equipped Recurrent Neural Network (RNN) for learning to determine and focus on reliable channels and temporal locations among the candidate multiple input sequences. $\operatorname{AttendMultiSource}$ produces re-weighted inputs, $\widehat{\mathbf X}$, based on the learned attention. This $\widehat{\mathbf X}$ is used for the next subnetwork $\operatorname{LSTM-AM}$, which is a Long Short-Term Memory (LSTM) acoustic model to estimate the probability of the output HMM state $\mathbf y$. Figure \ref{fig:schematic} visualizes our overall model with these two components. We describe more details of each component in the following subsections \ref{sec:attendmultisource} and \ref{sec:lstmam}.

\begin{align}
\hat{\mathbf X} &= \operatorname{AttendMultiSource} (\mathbf X, \mathbf y) \\
P(\mathbf y|\mathbf X) &= \operatorname{LSTM-AM} (\hat{\mathbf X}, \mathbf y) 
\end{align}

\subsection{Attention mechanism for multiple sources}
\label{sec:attendmultisource}
The challenge we attempt to address with the neural attention mechanism is the problem of misaligned multiple input sources with non-stationary quality over time. Specifically, in multi-channel distant speech recognition, the arrival time of each channel is different because the acoustic path length of each signal differs according to the location of the microphone. This results in the misalignment of input features. These differences in arrival time are even greater when the space between microphones is larger. Even worse, signal quality across channels can also vary over time because the speaker and interfering noise sources may keep changing. Figure \ref{fig:schematic} describes the asynchronous arrival of multiple inputs due to acoustic path length differences.

We now introduce an attention mechanism to cope with the misaligned input problem, and formulate the $\operatorname{AttendMultiSource}$. At every output step $t$, the $\operatorname{AttendMultiSource}$ function produces a re-weighted input representation $\widehat{\mathbf X_c}$, given $c$th candidate input set $\mathbf X_c$. $\mathbf X_c$ is a subsequence of time frames. As proposed by \citep{bahdanau2015end}, we perform similar windowing to limit the exploring temporal location of inputs for computational efficiency and scalability. We limit the range of attention to $l$=7 time frames ($\pm 3$). In our experiments, longer time steps had little impact on overall performance and would rather benefit from microphones placed further apart from each other. 

For re-weighting the input $\mathbf X_c$, $\operatorname{AttendMultiSource}$ predicts an attention weight matrix $\mathbf{A}^{time,ch}_{t}$ at each output step $t$. Unlike previous attention mechanisms, we produce a weight matrix rather than a vector, because our attention mechanism additionally identifies which channel, in a given time step, is more relevant. Therefore, $\mathbf{A}^{time,ch}_{t}$ is the (\textit{number of channels}) by (\textit{number of candidate input frames}) matrix - here it is $N$ x $l$ matrix. Attention weights are calculated based on four different information sources: 1) attention history $\mathbf{A}^{time,ch}_{t-1}$, 2) content in the candidate sequences $\mathbf X_c$, 3) decoding history $\mathbf s_{t-1}$, and 4) additional spatial information between multiple microphones based on phase difference information $\mathbf{PD}_c$ corresponding to $\mathbf X_c$. The following three formulations describe the $\operatorname{AttendMultiSource}$ function: 

\begin{align}
\mathbf{E}^{time,ch}_{t} &= \operatorname{MLP}(\mathbf{s}_{t-1}, \mathbf{A}^{time,ch}_{t-1}, \mathbf{PD}_c, \mathbf X_c) \label{E:eqn1}\\
\mathbf{A}^{time,ch}_{t} &= \operatorname{softmax}(\mathbf{E}^{time,ch}_{t}) \label{E:eqn2}\\
\widehat{\mathbf X_c} &= \mathbf{A}^{time,ch}_{t} \cdot \mathbf X_c \label{E:eqn3}
\end{align}

Specifically, $\operatorname{MLP}$ (in equation \ref{E:eqn1}) computes an energy matrix $\mathbf{E}^{time,ch}_{t}$ ($N$ x $l$) by the following equation: 

\begin{equation}
\mathbf{E}^{time,ch}_{t} = 
            tanh( \mathbf W_s \cdot \mathbf{s}_{t-1}
            + \mathbf W_a \cdot \mathbf{A}^{time,ch}_{t-1} 
            + \mathbf W_p \cdot \mathbf{PD}_c 
            + \mathbf W_x \cdot \mathbf X_c + b )
\end{equation}

where $\mathbf W_s$, $\mathbf W_a$, $\mathbf W_p$, and $\mathbf W_x$ are parameter matrices, and $b$ is a parameter vector. Once we compute the energy $\mathbf{E}^{time,ch}_{t}$ at time $t$, then we obtain  $\mathbf{A}^{time,ch}_{t}$ by normalizing $\operatorname{exp}(\mathbf{E}^{time,ch}_{t}) / \sum_{time,ch} \operatorname{exp}(\mathbf{E}^{time,ch}_{t}) $, such that, $\forall t$, $\mathbf{A}^{time,ch}_{t}$ $\geq 0$, and $\sum_{time,ch} {\mathbf{A}^{time,ch}_{t} = 1}$ (in equation \ref{E:eqn2}). Finally, re-weighted output $\widehat{\mathbf X_c}$ is generated by calculating the dot product of the attention weights $\mathbf{A}^{time,ch}_{t}$ and candidate input $\mathbf X_c$ (in equation \ref{E:eqn3}). Typically, the selection of elements from input candidates is a weighted sum. However, we only calculate the dot product in order to avoid losing information. 

To accelerate the learning of the attention mechanism, we use additional spatial information based on analysis of differences in arrival time. It is generally assumed that the human auditory system can localize multiple sounds and attend to the desired signal using information from the interaural time difference (ITD) \citep{stern2008binaural, park2009spatial}. A previous study \citep{kim2009signal} attempted to emulate human binaural processing and estimate ITD indirectly by comparing the phase difference between two microphones at each frequency domain. The authors identified a "close" time-frequency component to the speaker based on the estimated ITD. Similarly, we use the phase difference between two microphones to infer spatial information. The following equations are used to compute phase difference between two microphones $i$ and $j$, where $i \ne j, i,j \in \{1 \cdots N\}$:

\begin{align}
{pd}^{ch_i-ch_j} = \min |\angle x^{ch_i} - \angle x^{ch_j} - 2\pi r| \\
\mathbf{PD}^{ch_i-ch_j} = (pd^{ch_i-ch_j}_1, \cdots, pd^{ch_i-ch_j}_T) \\
\mathbf{PD} = \{\mathbf{PD}^{ch_1-ch_2}, \cdots, \mathbf{PD}^{ch_4-ch_5}\} 
\end{align}

From these equations, we calculate the phase differences of each time-frequency bin of each pair of multiple microphones. In our work, we use 256 frequency bins for 25ms windows. The phase feature $\mathbf{PD}$ is calculated in every pair of channels, then the $\operatorname{MLP}$ network accepts the $\mathbf{PD}_c$ corresponding to the input candidates, with $\mathbf{X}_c$ as an additional input. 

\subsection{LSTM Acoustic Model}
\label{sec:lstmam}
Our next subnetwork $\operatorname{LSTM-AM}$ serves as a typical RNN-based acoustic model, except that it accepts the re-weighted input $\widehat{\mathbf X_c}$ instead of the original input $\mathbf X_c$. $\operatorname{LSTM-AM}$ uses a Long Short-Term Memory RNN (LSTM)\citep{hochreiter1997long}, which has been successfully applied to speech recognition tasks due to its ability to handle long-term dependencies. The LSTM contains special units called memory blocks in the recurrent hidden layer, and each block has memory cells $c_t$ with special three-gates (input $i_t$, output $o_t$, and forget $f_t$) to control the flow of information. 

In our work, we use a simplified version of an LSTM without peephole connections and biases to reduce the computational expense of learning the standard LSTM models. Although LSTMs have many variations for enhancing their performance, such as BLSTM \citep{graves2013hybrid}, LSTMP \citep{sak2014long}, and PBLSTM \citep{chan2015listen}, in our work, we focus on verifying an additional attention mechanism with a simple LSTM architecture, instead of improving LSTM acoustic modeling overall.

$\operatorname{LSTM-AM}$ maps a re-weighted input sequence based on the attention mechanism  $\widehat{\mathbf X} = \{ \widehat{\mathbf x^{ch_1}}, \cdots, \widehat{\mathbf x^{ch_N}} \}$, where $\widehat{\mathbf x^{ch_i}} = (\widehat{{x^{ch_i}_1}}, \cdots, \widehat{{x^{ch_i}_T}})$, to an output sequence $\mathbf y_t = (y_1, \cdots, y_T)$ by calculating the network unit activations using the following equations iteratively from $t = 1$ to $T$: 

\begin{align}
{i_t} &= \sigma ( \widehat{\mathbf x_c} W_{xi} + {h_{t-1}} W_{hi} ) \label{E:eqn4}\\
{f_t} &= \sigma ( \widehat{\mathbf x_c} W_{xf} + {h_{t-1}} W_{hf} ) \label{E:eqn5}\\
{c_t} &= {f_t} \cdot c_{t-1} + {i_t} \cdot \tanh (\widehat{\mathbf x_c} W_{xc} + {h_{t-1}} W_{hc}) \label{E:eqn6}\\
{o_t} &= \sigma ( \widehat{\mathbf x_c} W_{xo} + {h_{t-1}} W_{ho} ) \label{E:eqn7}\\
{s_t} &= {o_t} \cdot \tanh ({c_t}) \label{E:eqn5}
\end{align}

where $W$ terms denote weight matrices, and $\sigma$ the logistic sigmoid function. $i_t$, $f_t$, $o_t$, and $c_t$ are the input gate, forget gate, output gate and cell activation vectors, respectively. Finally, the output $s_t$ is used to predict the current HMM state label by softmax (in equation \ref{E:eqn8}). $s_t$ is also used to predict the next $t+1$ attention matrix as well as the next $c_{t+1}$ hidden state of $\operatorname{LSTM-AM}$.

\begin{align}
y_t &= argmax_i P(y = i|s_t) \label{E:eqn8}
\end{align}

\section{Experiments}
\label{sec:exp}

\subsection{Dataset}
We evaluated the performance of our architecture on the CHiME-3 task. The CHiME-3 \citep{chime3} task is automatic speech recognition for a multi-microphone tablet device in an everyday environment - a cafe, a street junction, public transport, and a pedestrian area. There are two types of datasets: \texttt{REAL} and \texttt{SIMU}. The \texttt{REAL} data consists of 6-channel recordings. 12 US English speakers were asked to read the sentences from the WSJ0 corpus \citep{garofalo2007csr} while using the multi-microphone tablet. They were encouraged to adjust their reading positions, so that the target distance kept changing over time. The simulated data \texttt{SIMU} was generated by mixing clean utterances from WSJ0 into background recordings. To verify our method in a real noisy environment, we first chose not to use the simulated dataset but rather to use only the \texttt{REAL} dataset, with 5 channels from the five microphones, which were located in each corner of tablet, about 10cm to 20cm away from each other (we excluded one microphone, which faced backward in the tablet device). We then evaluated our system on the full CHiME3 dataset, \texttt{MULTI}, including \texttt{REAL} and \texttt{SIMU}. 

\subsection{System Training}
All the networks were trained on the 1,600 utterance (about 2.9 hours) \texttt{REAL} dataset and then on the 8,738 utterance (about 18 hours) \texttt{MULTI} dataset. The dataset was represented with 25ms frames of 40-dimensional log-filterbank energy features computed every 10ms. We produced 1,992 HMM state labels from a trained GMM-HMM system using near-field microphone data, and these state labels were used in all subsequent experiments. We use one layer of LSTM architecture with 512 cells. The weights in all the networks were initialized to the range (-0.03, 0.03) with a uniform distribution, and the initial attention weights were initialized to $1/n$ in $n$ dimensions. We set the configuration of the learning rate to 0.4 and after two epochs it decays during training. All models resulted in a stable convergence range from 1e-04 to 5e-04. To avoid the exploding gradient problem, we limited the norm of the gradient to 1 \citep{pascanu2012difficulty}. Apart from the gradient clipping, we did not limit the activations of the weights.

During training, we evaluated frame accuracies (i.e. phone state labeling accuracy of acoustic frames) on the development set of 1,640 utterances in \texttt{REAL} and 3,280 utterances in \texttt{MULTI}. The trained models were evaluated in a speech recognition system on a test set of 1,320 utterances. For all the decoding experiments, we used a size 18 beam and size 10 lattices. There is a mismatch between the Kaldi baseline \citep{Povey_ASRU2011} and our results because we did not perform sequence training (sMBR) or language model rescoring (5-gram rescoring or RNNLM). The inputs for all networks were log-filterbank features, with 5 channels stacking, and then with 7 frames stacking (+3-3). 

\begin{table}[t]
\caption{Comparison of WERs(\%) on development and evaluation set of the subset (REAL) of the CHiME-3 task between the three baseline systems, and our proposed framework, ALSTM. The models are trained on on real data (3hrs).}
\label{tab:res}
\begin{center}
\begin{tabular}{l||c|c}
\multicolumn{1}{c}{\bf MODEL (Input)}  &\multicolumn{1}{c}{\bf DEV (WER \%)} &\multicolumn{1}{c}{\bf TEST (WER \%)}
\\\hline
\hline
\textit{Baselines - Real Data (3hrs)} & & \\
LSTM (Preprocessing 5 noisy-channel) & 35.2 & 52.1 \\
LSTM (single noisy-channel)   & 39.1 & 57.1  \\
LSTM (5 noisy-channel)  & 43.0 & 60.1  \\
\hline
\textit{Proposed - Real Data (3hrs)} & & \\
ALSTM    & 35.9 & 52.3  \\
ALSTM (with phase) & 33.9 &	50.0  \\
\hline

\end{tabular}
\end{center}
\end{table}

\begin{table}[t]
\caption{Comparison of WERs(\%) on development and evaluation set of the subset (REAL) of the CHiME-3 task between the baseline system, and our proposed framework, ALSTM. The models are trained on on real + simulated data (18hrs).}
\label{tab:res2}
\begin{center}
\begin{tabular}{l||c|c}
\multicolumn{1}{c}{\bf MODEL (Input)}  &\multicolumn{1}{c}{\bf DEV (WER \%)} &\multicolumn{1}{c}{\bf TEST (WER \%)}
\\\hline
\hline
\textit{Baselines - Real + Simulated Data (18hrs)} & & \\
LSTM (Preprocessing 5 noisy-channel) & 18.6	& 32.0 \\
\hline
\textit{Proposed - Real + Simulated Data (18hrs)} & & \\
ALSTM (with phase)   & 16.5	& 26.5 \\
\hline

\end{tabular}
\end{center}
\end{table}
\subsection{Results}
In Table \ref{tab:res} and \ref{tab:res2}, we summarize word error rates (WERs) obtained on the subset of the CHiME3 task. ALSTM is our proposed model, which has an attention mechanism for multiple inputs as described in \ref{sec:attendmultisource}, and ALSTM (with phase) used phase information in addition to ALSTM. 

As our baselines, we built three models on the \texttt{REAL} dataset and used the same simple version of the LSTM architecture that we described in Section \ref{sec:lstmam} with three different inputs. LSTM (Preprocessing 5 noisy-channel) was trained on the enhanced signal from 5 noisy channels. We obtained the enhanced signal from the beamforming toolkit, which was provided by the CHiME3 organizer \citep{chime3,loesch2010adaptive,blandin2012multi,mestre2003diagonal}. LSTM (single noisy-channel) was trained on a single noisy channel, and LSTM (5 noisy-channels) used the concatenated 5 noisy channels. We also built LSTM (Preprocessing 5 noisy-channel) on the \texttt{MULTI} dataset.   

As expected, LSTM (Preprocessing 5 noisy-channel) provided a substantial improvement in WER compared to LSTM (single noisy-channel) and LSTM (5 noisy-channel), showing a 13.3\% and 5.0\% relative improvement in WER, respectively. We also found that the model, which simply combined 5 features across microphones, did not perform very well. It showed poorer results than even the model trained with single microphone data. This result underscores the importance of integrating channels based on analysis of differences in arrival times.

Our model with the attention mechanism provided a significant improvement in WER compared to LSTM (5 noisy-channel). Compared to LSTM (5 noisy-channel), ALSTM (with phase) achieved a 17\% reduction in relative error rate on the evaluation set, and ALSTM achieved a 13\% relative error rate. These results suggest that we can leverage the attention mechanism to integrate multiple channels efficiently. To ensure the improvement of the system was coming from our time-channel attention mechanism, we compared our model to a model with an attention mechanism across time only on single-channel input. This comparison model helped to improve accuracy by 3\%, a lower gain than that achieved by the time-channel attention mechanism. 

We also found that the additional phase information can help to learn attention and WER improved by 4.6\% relatively. In comparison with LSTM (Preprocessing 5 noisy-channel), we found that our proposed model achieved comparable performance to beamforming without any preprocessing. Although ALSTM shows a slightly lower performance as compared to LSTM (Preprocessing 5 noisy-channel), a 4.0\% relative error rate was obtained by ALSTM (with phase). When we used LSTM-AM with the additional phase features without any attention mechanism, it had a negative influence on learning. Thus, using the phase features for the attention mechanism is more effective than using the phase features as direct inputs of the acoustic model.

We also evaluated the models on the \texttt{MULTI} dataset. We found that our system outperformed LSTM (Preprocessing 5 noisy-channel) by 5\%, and the gain from the time-channel attention mechanism increased.  

We then analyzed the computational aspects of our system. As the multi-microphone is performed as part of the acoustic model computation we have actually found it to be more computationally efficient than performing beamforming followed by an LSTM acoustic model. On our development machine (Intel(R) Xeon(R) CPU E5-2690 v3 @ 2.60GHz), the proposed multi-microphone model with attention and phase operated 0.08 real-time, which was significantly faster than the beamforming followed by acoustic model computation which operated at 0.6 real-time. 


\section{Conclusions}
\label{sec:conclusion}
We proposed an attention-based model (ALSTM) that uses asynchronous and non-stationary inputs from multiple channels to generate outputs. For a distant speech recognition task, we embedded a novel attention mechanism within a RNN-based acoustic model to automatically tune its attention to a more reliable input source. We presented our results on the CHiME3 task and found that ALSTM showed a substantial improvement in WER. Our model achieved comparable performance to beamforming without any prior knowledge of the microphone layout or any explicit preprocessing.

The implications of this work are significant and far-reaching. Our work suggests a way to build a more efficient ASR system by bypassing preprocessing. Our findings suggest that this approach will likely do well on tasks that need to exploit misaligned and non-stationary inputs from multiple sources, such as multimodal problems and sensory fusion. We believe that our attention framework can greatly improve these tasks by maximizing the benefits of using inputs from multiple sources. 

\subsubsection*{Acknowledgments}
The authors would like to acknowledge Richard M. Stern and William Chan for their valuable and constructive suggestions. This research was supported by LGE.

\bibliography{iclr2016_conference}
\bibliographystyle{iclr2016_conference}

\end{document}